\documentclass[letterpaper, 10 pt, journal, twoside]{IEEEtran}

\IEEEoverridecommandlockouts                              

\usepackage{amsmath} 
\usepackage{amssymb}  

\usepackage{hyperref}
\usepackage{graphicx}
\usepackage{cleveref}
\usepackage[export]{adjustbox}
\usepackage[skip=1pt]{subcaption}
\usepackage{multirow}
\usepackage{algorithm}
\usepackage{algpseudocode}

\usepackage{tikz}
\newcommand*\circled[1]{\tikz[baseline=(char.base)]{\node[shape=circle,draw,inner sep=1pt] (char) {#1};}} 

\usepackage[font=small]{caption}

\title{
Sparse-Dense Motion Modelling and Tracking for Manipulation without Prior Object Models
}

\author{Christian Rauch$^{1}$, Ran Long$^{1}$, Vladimir Ivan$^{1}$, Sethu Vijayakumar$^{1}$
\thanks{Manuscript received: December, 10, 2021; Revised February, 18, 2022; Accepted April, 11, 2022.}%
\thanks{This paper was recommended for publication by Editor Markus Vincze upon evaluation of the Associate Editor and Reviewers' comments.
This research is supported by the EU H2020 project Enhancing Healthcare with Assistive Robotic Mobile Manipulation (HARMONY, 9911237), The Alan Turing Institute and the Kawada Robotics Corporation.}%
\thanks{$^{1}$School of Informatics, University of Edinburgh, Edinburgh EH8 9AB, U.K. {\tt\small Christian.Rauch@ed.ac.uk}}%
\thanks{Digital Object Identifier (DOI): see top of this page.}%
}

\begin{document}
\maketitle

\begin{abstract}

This work presents an approach for modelling and tracking previously unseen objects for robotic grasping tasks. Using the motion of objects in a scene, our approach segments rigid entities from the scene and continuously tracks them to create a dense and sparse model of the object and the environment. While the dense tracking enables interaction with these models, the sparse tracking makes this robust against fast movements and allows to redetect already modelled objects.

The evaluation on a dual-arm grasping task demonstrates that our approach 1) enables a robot to detect new objects online without a prior model and to grasp these objects using only a simple parameterisable geometric representation, and 2) is much more robust compared to the state of the art methods.

\end{abstract}


\section{Introduction}

\IEEEPARstart{R}{obotic} grasping tasks typically require a detailed visual and geometric representation of all target objects that the robot might interact with. These detailed representations are provided classically as textured mesh model \cite{Pauwels2014, Schmidt2015a} or in form of a trained segmentation or pose estimation approach \cite{Schmidt2017, Ruenz2018}. In constrained environments with known and fixed sets of objects, these approaches have proven very efficient, which is supported by the large corpus of work in this area. 

However, these approaches do not scale well with a growing number of objects. New models have to be created tediously by manually defining the geometric shape and the texture, or by manually labelling training data. Additionally, increasing the set of possible objects that the robot might encounter in a new scenario adds unnecessary redundancy and computational costs when the actually encountered objects only make up a fraction of the entire dataset.

We argue that instead of providing such specific object models, we can acquire objects-of-interest online during a particular task and thus create a task-specific set of target objects that the robot can interact with.

In a robotic grasping application without prior models, we have to consider a couple of additional challenges when tracking and modelling objects online:
\begin{itemize}
    \item without an \textit{a-priori} model we cannot rely on a given reference frame as grasp target,
    \item the limited camera field-of-view restricts the long-term tracking of objects,
    \item to expand the field-of-view (FoV), tracking must handle large and fast view-point changes,
    \item to prevent the system from modelling each new object separately, we must redetect previously and partially reconstructed objects.
\end{itemize}

The problem of modelling and tracking borrows tools from Simultaneous Localisation and Mapping (SLAM) and applies those to multiple rigid entities in a scene, including the environment itself.
Dense SLAM methods with a dense Iterative Closest Points (ICP) loss function typically suffer from correspondence ambiguity between consecutive point clouds. This restricts their application to slow motions or high sensor update rates in combination with fast per-image processing. Sparse methods with distinct keypoints on the other hand provide more robust correspondences, but rely on very accurate keypoint localisation and do not provide a dense model as required by many robotic applications, such as navigation and obstacle avoidance or manipulation tasks.

To overcome these limitations and enable a robot to build a task-specific set of target objects online, we propose a combination of dense and sparse tracking methods that directly use the dense and sparse visual motion cues to robustly track and densely model moving objects. To prevent the repetitive modelling of previously seen objects, we further facilitate the robustness of sparse features for redetecting partial models for long-term modelling. In summary, this work contributes a model-free tracking method for robotic grasping tasks that:
\begin{enumerate}
    \item robustly initialises tracking between consecutive image frames to handle fast view-point changes,
    \item segments objects of interest directly by visual motion cues instead of relying on geometric differences, and
    \item redetects previously seen and partially modelled objects to reuse information for long-term modelling.
\end{enumerate}

\section{Related Work}

\subsection{Model-Based Tracking}
While we are predominantly interested in model-free tracking without an \textit{a-priori} model, we also borrow methods from model-based approaches once an initial partial model has been established. Classic 3D pose estimation and tracking approaches rely on a geometric \cite{Schmidt2015a} and visual \cite{Pauwels2014} representation of the target model in 3D. These models provide the gold-standard reference for comparing the geometric and visual features with the actual sensor data for the tracking loss function. More recent tracking methods directly incorporate the tracking loss in a deep Convolutional Neural Network (CNN) and formulate tracking as a classification or regression problem. Such semantic tracking methods initially use a labelled training set of the target objects and eventually represents them in a latent space \cite{Schmidt2017,Ruenz2018}. Hybrid methods can use the geometric and visual model to generate such training data \cite{Rauch2019}.

These mesh and semantic model-based approaches have in common that the information about objects has to be collected manually and given \textit{a-priori}.
This bias towards user-chosen object models does not scale to new scenarios and limits the applicability by either providing too many or too few object models to cover all potential objects or to provide an efficient coverage, respectively.

\subsection{Single and Multiple Transformation Estimation}
SLAM methods are inherently model-free and focus on localisation with respect to the environment, assuming it as the only visible rigid entity in the scene \cite{Whelan2015,Mur-Artal2015}. Similarly, object modelling approaches use the same techniques, but usually only focus on a single model at a time\cite{Krainin2011, Weise2011, Tzionas2015, Wang2019}. This single-model assumption does not hold in many practical applications where we have to deal with additional motion.
While ElasticFusion \cite{Whelan2015} accounts for drift via loop closure, it does not handle additional motion explicitly. StaticFusion \cite{Scona2018} explicitly segments and neglects dynamic objects as outliers by adapting a reprojection error threshold online. Co-Fusion \cite{Ruenz2017} and RigidFusion \cite{Long2021} further extend these approaches by explicitly tracking additional rigid transformations to model additional motion, with RigidFusion additionally relying on kinematic motion priors.

Tracking multiple objects in parallel without an \textit{a-priori} model encompasses the need to segment a scene based on the discrepancy of model transformations.
Ideally, the residual of the transformation estimation loss function can be used as a metric for associating data to transformations, but we will show that this approach is not reliable in some cases and propose an alternative metric.

\subsection{Dense and Sparse Representation}
In parallel tracking and modelling approaches, the model representation and the tracking loss function are tightly coupled. This representation varies from sparse keypoints \cite{Judd2018} to dense point clouds \cite{Ruenz2017}. While a dense reconstruction provides the highest quality of the model for robotic applications, using the distance between raw points as loss function leads to ambiguity and many local minima. This ambiguity requires that consecutive frames are close and thus limits the velocity of the motion. Sparse points can encode much more visual information from around a point's neighbourhood to create distinguishable points. Depending on how distinct the encoded information is, the points can be used for short-distance odometry \cite{Kitt2010} or longer-distance point matching \cite{DeTone2018}. Dense feature points ideally provide such distinct points densely over the image \cite{Schmidt2017,Florence2018}, but only have been demonstrated so far on object-specific datasets and thus cannot be applied in a model-free manner.

In aiming at combining the robustness of distinct keypoints with the dense model representation as required by robotic grasping tasks, we propose a combination of dense and sparse representations with the ability to directly segment objects of interest via visual motion cues and the ability to redetect previously seen objects for long-term tracking.

\section{Methodology}

\subsection{Problem Formulation}

RGB-D multi-motion tracking operates on a continuous stream of intensity $\mathbf{I}: \mathbb{R}^{W \times H} \mapsto \mathbb{R}$ and depth $\mathbf{D}: \mathbb{R}^{W \times H} \mapsto \mathbb{R}$ image pairs $(\mathbf{I}, \mathbf{D})^t$ representing the scene $\mathcal{S}$ at a certain point in time $t$.
A 3D point $\mathbf{p}$ in the camera frame is projected onto the 2D image plane as coordinate $\mathbf{x}$ using the pinhole projection $\pi: \mathbb{R}^3 \mapsto \mathbb{R}^2 \times \mathbb{R}$.
Chaining the projection and a 3D transformation, we can formulate the rigid warp field $\omega: \mathbb{R}^2 \mapsto \mathbb{R}^2$ with $\omega(\mathbf{x},\mathbf{T}) = \pi(\mathbf{T}\pi^{-1}(\mathbf{x},\mathbf{D}(\mathbf{x})))$.

\begin{figure}
    \centering
    \includegraphics[width=\linewidth]{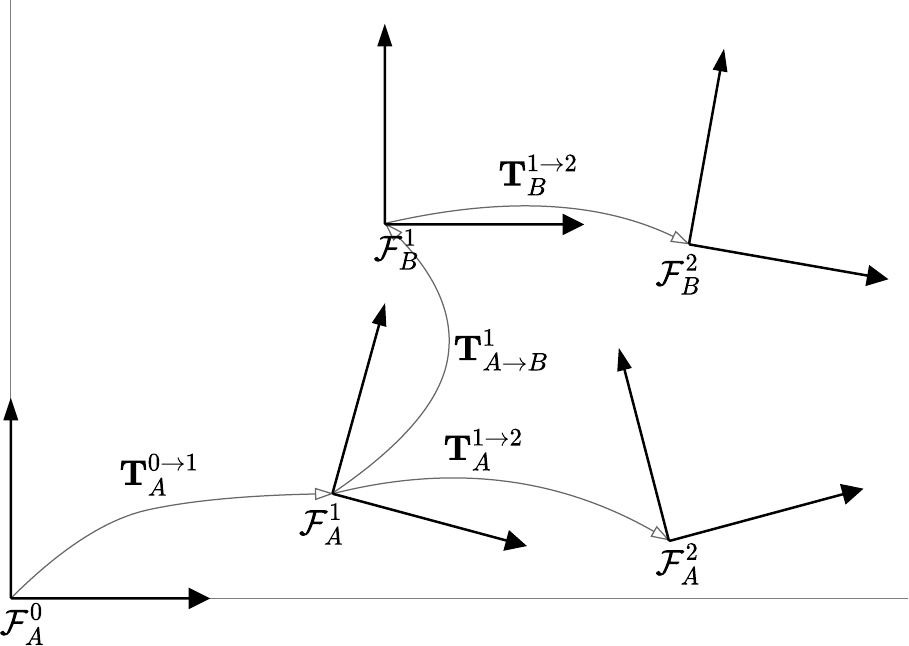}
    \caption{Transformation between frames. The motion of an object $A$ between time $0$ and $1$ is described as the transformation $\mathbf{T}_A^{0 \rightarrow 1}$ between the frames $\mathcal{F}_A^0$ and $\mathcal{F}_A^1$. The spatial relation between two objects $A$ and $B$ at time $1$ is given by the transformation $\mathbf{T}_{A \rightarrow B}^{1}$ between the frames $\mathcal{F}_A^1$ and $\mathcal{F}_B^1$. The initial frame $\mathcal{F}_A^0$ is defined as the world frame.}
    \label{fig:frames}
\end{figure}

The aim of multi-motion tracking and segmentation is to estimate the pose and visual representation of all $M$ moving entities $\mathcal{O}$ in a scene. At every point in time $t$, the scene can then be represented as a set $\mathcal{S}: \{\mathcal{O}_i\,|\, 0 \leq\: i < M\}$ of objects $\mathcal{O}: (\mathbf{T}, \mathcal{R})$ with their current pose $\mathbf{T}\in SE(3)$ and a 3D representation $\mathcal{R}$. Each $\mathcal{R}$ is defined within a coordinate frame $\mathcal{F}_i^t$ for a specific object $i$ and point in time $t$ (\Cref{fig:frames}).
Tracking provides the pose of frames $\mathcal{F}^t$ over time by estimating $\mathbf{T}^{(t) \rightarrow (t+1)}$ between these frames, and the segmentation provides the number of frames $\mathcal{F}_i$ at one point in time. At $t=0$, we assume that all initially observed data belongs to $\mathcal{R}$ of the first object frame $\mathcal{F}_{i=0}^{t=0}$ and define the first frame as the environment $\mathcal{S}: \{\mathcal{O}_0\}$. All consecutive object frames $\mathcal{F}_{i>0}$ are then spawned from the frame in which a new motion segment is detected.

All objects in $\mathcal{S}$ are represented by a combination of dense and sparse 3D points as $\mathcal{R}: \left(\mathcal{P}, \mathcal{K}\right)$. The dense representation is an unordered point cloud $\mathcal{P}: \{\mathbf{p}_i\,|\, 0 \leq\: i < N_p\}$ with $N_p$ points $\mathbf{p}\in\mathbb{R}^3$. The sparse representation is a set $\mathcal{K}: \{k_i\,|\, 0 \leq\: i < N_k\}$ of $N_k$ keypoints $k: (\mathbf{p}, \mathbf{f})$ with 3D coordinate $\mathbf{p}\in\mathbb{R}^3$ and feature vector $\mathbf{f}\in\mathbb{R}^{256}$. An example environment point cloud $\mathcal{P}$ is visualised in \Cref{fig:kpinit_reconstr,fig:kpinit_reconstr_error}.

The segmentation $\mathbf{S}: \mathbb{R}^{W \times H} \mapsto \mathbb{N}$ is defined in the image frame and associates each pixel $\mathbf{x}\in\mathbb{R}^2$ to a segment $s$ if $\mathcal{O}_i$ is visible at that pixel. At $t=0$ we assume that only $\mathcal{O}_0$ is visible, hence $\mathbf{S}(\mathbf{x}) = 0 \, \forall \, \mathbf{x} \in \mathbb{R}^{W \times H}$.

In summary, we are looking for a set of rigid objects whose union of individual rigid motions and visual representation explains all motion in the currently observed scene:
\begin{align}
    \mathop{\arg\min}_{\{(\mathbf{T}, \mathcal{R})\}} \sum_m^M \sum_\mathbf{x}^{\mathbb{R}^{W \times H}} \left\vert \mathcal{R}^{t}(\mathbf{x}) - \mathbf{T}_{(m)}^{(t-1) \rightarrow (t)}\mathcal{R}_{(m)}^{t-1}(\mathbf{x}) \right\vert \quad .
\end{align}
That is, given the current representation of the scene, as observed by a RGB-D camera, we are looking for a set of transformations $\{\mathbf{T}\}$ that, when applied to a corresponding set of visual representations $\{\mathcal{R}\}$, reconstructs the currently observed scene representation. The problem in model-free tracking is that we neither know how many objects or transformations $M$ exist nor do we know their full representation or which pixel $\mathbf{x}$ belongs to which object $m$.

\subsection{Overview}

\begin{figure}
    \centering
    \includegraphics[width=\linewidth]{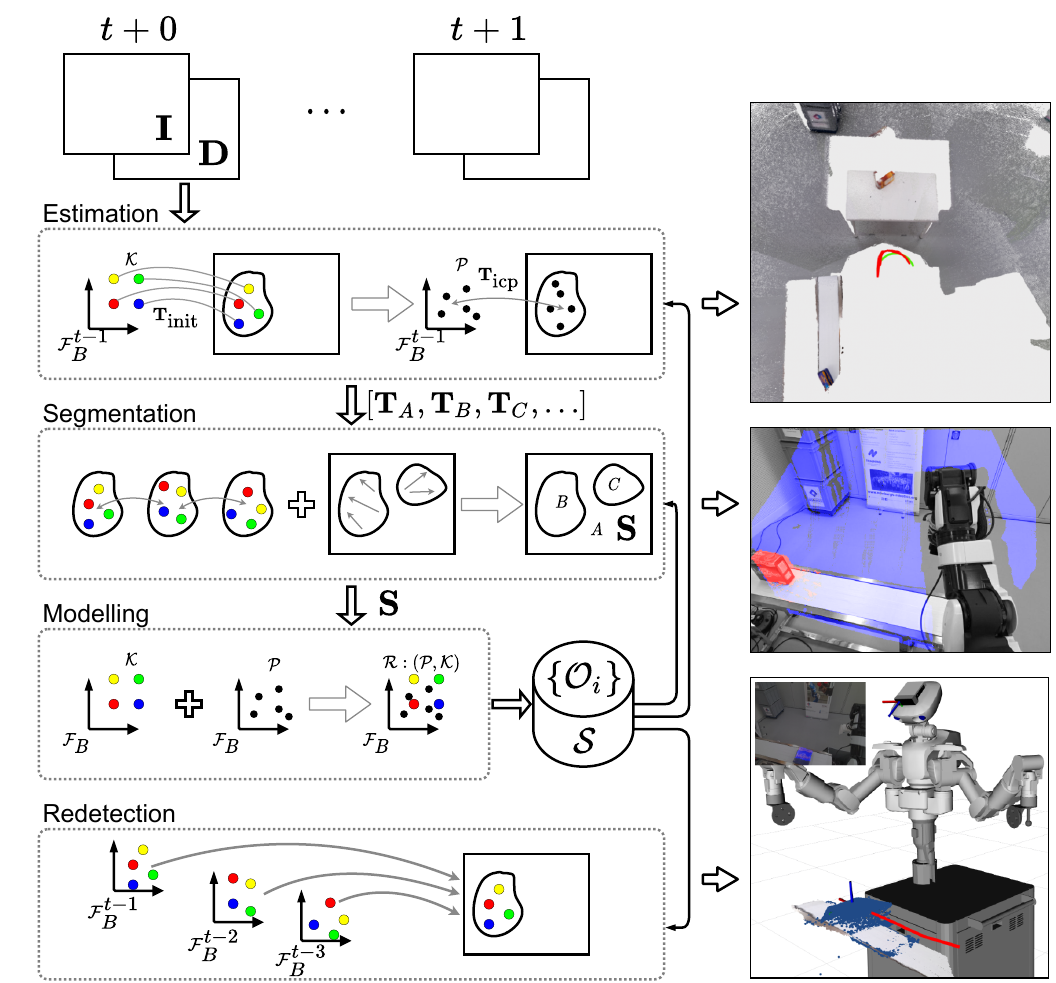}
    \caption{Multi-motion estimation and segmentation pipeline. From the previous or initial set of objects in $\mathcal{S}$, we initially estimate all transformations via keypoint correspondences followed by a dense refinement via ICP. The estimated transformations $\mathbf{T}$ provide the keypoint reprojection error that is seeding the optical flow segmentation $\mathbf{S}$. The model representation $\mathcal{R}$ is created by registering the image data in $\mathbf{S}$ using $\mathbf{T}$. The history of all $\mathcal{R}$ is compared to the current segmentation to detect previously seen models.}
    \label{fig:overview}
\end{figure}

The proposed approach operates in four consecutive phases on the individual image pairs to estimate the trajectory of individually moving object frames (\Cref{fig:overview}):
\begin{enumerate}
    \item \textbf{Estimation:} The sparse keypoints $\mathcal{K}$ of each model from the model database $\mathcal{S}$ are associated to the keypoints of the current image to estimate an initial transformation $\mathbf{T}_{\text{init}}$ via RANSAC (RANdom Sample Consensus). This transformation is used to initialise a dense ICP method on the dense point cloud $\mathcal{P}$, to refine this transformation as $\mathbf{T}_{\text{icp}}$ on all visible depth data and all tracked models.
    \item \textbf{Segmentation:} The sparse reprojection error from the estimated transformation on the last keypoint tracks and the optical flow on the colour image is used in a CRF (Conditional Random Field) to densely associate pixels to models in the current segmentation $\mathbf{S}$.
    \item \textbf{Modelling:} The segmented sparse and dense points are registered via the transformation $\mathbf{T}$ into the reference frame of each model. This provides the time-indexed visual model representation $\mathcal{R}$.
    \item \textbf{Redetection}: The time-indexed $\mathcal{K}$ for each inactive (not tracked) model is compared to the currently segmented keypoint sets to determine if a model is visible again, in which case the previously inactive model will be tracked and modelled again.
\end{enumerate}

For the aim of providing an object representation that can be used in typical robotic applications such as navigation and manipulation, we are primarily interested in dense representations. While dense depth data provides a sufficient amount of raw 3D information for this purpose, it does not have sufficient discriminative information to distinguish between different points or associate them directly. This ambiguity is a major limitation when operating at high velocities, estimating motion, and when trying to associate separate point clouds with each other, such as when determining if an object is already present in the scene. For these reasons, we propose to represent the object via sparse discriminative keypoints and dense raw points and use this representation throughout the estimation, segmentation and redetection phase.

\subsection{Transformation Estimation}

For each new colour and depth image pair, we first estimate the transformation between each object's reference frame to the camera frame using the sparse keypoints for an initial transformation and the dense point cloud for the refinement. This model-to-frame alignment thus uses the history of the object's sparse and dense representation and aligns this with the currently observed scene representation.

\subsubsection{Sparse Estimation}

The keypoints are extracted from $\mathbf{I}$ using SuperPoint \cite{DeTone2018}. Given the grey-scale image $\mathbf{I} \in \mathbb{R}^{W \times H}$, the core keypoint network encodes the image into a compressed representation $\mathcal{B} \in \mathbb{R}^{W/8 \times H/8 \times 128}$ which is then further processed by two branches and upscaled to a heatmap $\mathcal{X} \in \mathbb{R}^{W \times H}$ and a featuremap $\mathcal{D} \in \mathbb{R}^{W \times H \times 256}$. The heatmap represents the probability of a keypoint on that pixel coordinate, while the featuremap provides the $256$-dimensional normalised feature vector for that pixel coordinate. To reduce the amount of keypoints that are direct neighbours, we apply non-maximum suppression by maximum pooling in a $3 \times 3$ neighbourhood and also remove all responses for values below $0.015$. The keypoint set is then
\begin{align}
    \mathcal{K}: \{(\mathbf{p}, \mathbf{f}) \, | \, \mathbf{p} &= \pi^{-1}(\mathbf{x}, \mathbf{D}(\mathbf{x})),\\\nonumber
    \mathbf{f} &= \mathcal{D}(\mathbf{x}), \\\nonumber
    &\forall \, \mathbf{x} \ni \mathcal{X}(\mathbf{x}) > 0.015 \} \quad ,
\end{align}
where $\mathbf{p}$ denotes the back-projected keypoint coordinate in the camera frame and $\mathbf{f}$ denotes the feature vector.

We are looking for a transformation $\mathbf{T}_{(m)\text{init}}^{(t-1) \rightarrow (t)}$ of the $m$-th model that minimises the average distance of the 3D coordinates $\mathbf{p}$ and the 256D feature vectors $\mathbf{f}$ from keypoints $\mathcal{K}^{t-1}$ of the previous image to the keypoints $\mathcal{K}^t$ of the current image. This is done in two stages: First, we exhaustively search for $N_{m}$ keypoint correspondences $\{(u,v)_i \, | \, 0 \leq\: i < N_{m}\}$, that individually minimise the feature vector distance between keypoints in the model and the current image,
\begin{align}
    \mathop{\arg\min}_{(u,v)} \lVert \mathbf{f}_u^{t-1} - \mathcal{D}^{t}(v) \rVert^2 , \label{eq:loss_feat}
\end{align}
s.t. coordinates $u \in \mathbb{R}^2$ and $v \in \mathbb{R}^2$ are associated only if their feature vectors have mutually the closest distance in the set of possible one-to-many matches. Here, $\mathbf{f}_u$ denotes the feature vector of the last model keypoint that was originally observed at $u$.
Second, with the the minimised feature vector distances, we now minimise the sparse transformation loss:
\begin{align}
    \mathop{\arg\min}_{\mathbf{T}_{\text{init}}} \sum_i^{\lvert \mathcal{K} \rvert} \left\Vert \mathbf{p}_{u_i}^{t-1} - \mathbf{T}_{(m)\text{init}}^{(t-1) \rightarrow (t)} \mathbf{p}_{v_i}^{t} \right\Vert^2 \label{eq:est_transf_sparse}
\end{align}
where $ \mathbf{p}_{u_i}^{t-1}$ denotes the keypoint coordinate in the model's reference frame from $\mathcal{K}$, originally observed at $u$, and $\mathbf{p}_{v_i}^{t}$ denotes the back-projection of correspondence $v$ into the camera frame. Note that the correspondences $(u,v)$ are established between the representation $\mathcal{K}$ of an object model and the current image frame, not between consecutive image frames. This reduces outliers within the correspondences but might still retain wrongly associated keypoints due to ambiguity in the feature space and an inaccurate segmentation. To robustly estimate the model-to-frame transformation that minimises \labelcref{eq:est_transf_sparse}, we apply RANSAC to repeatedly sample possible inliers from the set of correspondences, least-squares optimise \labelcref{eq:est_transf_sparse} on the inliers subset, and finally select the transformation that produces the lowest error with a minimum set of inliers. This estimation provides an initial sparse transformation $\mathbf{T}_{\text{init}}$ between each model's reference frame and the camera frame.

\subsubsection{Dense Estimation}

The sparse estimation only considers a very small amount of data from the model representation $\mathcal{K}$, that might additionally not be equally distributed and affected by quantisation errors of the pixel coordinates. To mitigate such effects, we propose to refine the initial sparse transformation using the raw dense data $\mathcal{P}$, that provides a much wider coverage. For this stage of the pipeline, we rely on the dense ICP implementation of Co-Fusion \cite{Ruenz2017}. The dense transformation loss is formulated similar to \labelcref{eq:est_transf_sparse} as the plane-to-point loss:
\begin{align}
    \mathop{\arg\min}_{\mathbf{T}_{\text{icp}}} \sum_\mathbf{x}^{\mathbb{R}^{W \times H}} \left(\left( \mathbf{p}_{\mathbf{x}}^{t-1} - \mathbf{T}_{(m)\text{icp}}^{(t-1) \rightarrow (t)} \mathbf{p}_{\mathbf{x}}^{t} \right) \cdot \mathbf{n}_{\mathbf{x}}^{t-1}\right)^2 \label{eq:est_transf_dense}
\end{align}
with $\mathbf{p}_{\mathbf{x}}^{t-1}$ and $\mathbf{n}_{\mathbf{x}}^{t-1}$ as the 3D coordinate and the normal, respectively, at pixel coordinate $\mathbf{x}$ of the dense object model transformed into the camera frame and projected onto the image plane. As before, the current point coordinates $\mathbf{p}_{\mathbf{x}}^{t}$ are obtained from the back-projection of $\mathbf{x}$ in the camera frame. Additionally to this depth derived loss, we also use the same colour derived loss from the baseline ICP implementation \cite{Ruenz2017}. In contrast to the sparse problem, this dense problem is solved using an iterative gradient-based approach.

The ambiguity of raw depth data leads to many local minima in this loss function and the optimisation thus has to be initialised close to the solution. Assuming low object motion or high camera sample rate, $\mathbf{T}_{\text{icp}}$ can be initialised at identity. To avoid local minima, we propose to initialise $\mathbf{T}_{\text{icp}}$ via the previously obtained sparse transformation $\mathbf{T}_{\text{init}}$ by pre-transforming the dense model representation $\mathcal{P}$ with $\mathbf{T}_{\text{init}}$. The optimisation for the pre-transformed dense loss,
\begin{align}
    \mathop{\arg\min}_{\mathbf{T}_{\text{icp}^*}} \sum_\mathbf{x}^{\mathbb{R}^{W \times H}} \left(\left( \mathbf{T}_{(m)\text{init}}^{-1}\mathbf{p}_{\mathbf{x}}^{t-1} - \mathbf{T}_{(m)\text{icp}^*}^{(t-1) \rightarrow (t)} \mathbf{p}_{\mathbf{x}}^{t} \right) \cdot \mathbf{n}_{\mathbf{x}}^{t-1}\right)^2 ,
\end{align}
is then also initialised at identity. The original transformation between the original model $m$ at $t-1$ and the camera frame at $t$ is then obtained by $\mathbf{T}_{(m)}^{(t-1) \rightarrow (t)} = \mathbf{T}_{(m)\text{init}}^{(t-1) \rightarrow (t)} \mathbf{T}_{(m)\text{icp}^*}^{(t-1) \rightarrow (t)}$.

\subsection{Segmentation}

While the estimation stage operates on a fixed model representation $\mathcal{R}$ for a fixed model set $\mathcal{S}$, the aim of the segmentation stage is to extend the set of known models and their visual representation, if necessary, to explain all motions in the scene.

One cue of motion is the sparse \labelcref{eq:est_transf_sparse} and dense \labelcref{eq:est_transf_dense} reprojection error represented per pixel. The error signifies how well a given transformation describes the observed motion in a scene, assuming low errors belong to inliers and high errors belong to outliers. As argued for the estimation before, this assumption only holds if there is no ambiguity in the data. Similarly to the local minima in the dense estimation, the ambiguity in the raw depth and colour data leads to ``false negatives'', where the reprojection error is low when the object is indeed moving. A trivial example of this effect is image plane parallel motion where the depth distance to the image plane does not change.

To circumvent this issue, we propose to rely on the keypoint reprojection error as the main cue of motion and propagate this cue to nearby pixels using optical flow. We formulate this as a Dense Conditional Random Fields (CRF) problem \cite{Kraehenbuehl2013}:
\begin{align}
    \sum_i \psi_i(s_i\,|\,\mathbf{\theta}) + \sum_{i<j} \psi_{ij}(s_i,s_j\,|\,\mathbf{\theta}) \label{eq:crf}
\end{align}
with the keypoint 2D reprojection drift as the unary potential
\begin{align}
    \psi_i(s_i\,|\,\mathbf{\theta}) = \frac{1}{\Delta t} \left\Vert \pi \left( \mathbf{p}_{u_i}^{t-1} \right) - \pi \left( \mathbf{T}_{(m)\text{init}}^{(t-1) \rightarrow (t)} \mathbf{p}_{v_i}^{t} \right) \right\Vert^2
\end{align}
and the pairwise potential
\begin{align}
    \psi_{ij}(s_i,s_j\,|\,\mathbf{\theta}) = \sum \mathbf{1}_{[s_i \neq s_j]} g(\mathbf{f}_i-\mathbf{f}_j)
\end{align}
with $g$ as a Gaussian kernel with diagonal covariance matrix of the feature space. The 4D feature space is defined by the 2D coordinate of a pixel $\mathbf{x}$ and its optical flow displacement vector between consecutive images $\mathbf{d}$, hence $\mathbf{f} = [\mathbf{x},\mathbf{d}]$. We use the drift $\Delta t = (t) - (t-1)$ of a keypoint between the previous and current image instead of the distance to account for irregular framerates. While the unary potential alone defines the probability that a keypoint belongs to one of the tracked transformations, the pairwise potential forces pixels with similar optical flow in the neighbourhood of a keypoint to be assigned to the same transformation, hence propagating the motion cue from the 2D keypoint reprojection error towards the dense neighbourhood using the optical flow.

This CRF only provides a reasonable solution when there is motion in the scene. To handle low motion and static objects, we weight the motion probability with the flow magnitude $\lVert \mathbf{d} \rVert^2$ and combine it with the probability inferred from the dense reprojection in \labelcref{eq:est_transf_dense}.

\subsection{Modelling and Redetection}

Given all objects' estimated poses and corresponding segments over time, we can transform all $\mathcal{R}_{(m)}^t$ via $\mathbf{T}_{(m)}^t$ and register them as $\mathcal{R}_{(m)}$ into a common reference frame.

When the keypoint initialisation \labelcref{eq:est_transf_sparse} fails or the object segment \labelcref{eq:crf} is too small, an object is flagged as lost and moved from $\mathcal{S}$ to a new set $\mathcal{S}_\text{lost}$ of untracked objects to prevent model corruption in case of tracking failures.

The history of these lost objects' $\mathcal{K}$ is continuously matched to the current image keypoints, segmented into rigid parts via $\mathbf{S}_t$ (\Cref{algo:detect}). The limited in-plane rotational invariance of the feature descriptors \labelcref{eq:loss_feat} makes it necessary to store and match the entire history of $\mathcal{K}$. Because we match over all models and all their history, the runtime of the redetection grows over time.

After an object has been redetected, its currently tracked duplicate is removed and the original object model is moved back to $\mathcal{S}$ with the new pose. From thereon, the object is again part of the regular tracking and modelling pipeline.

\begin{algorithm}
\begin{algorithmic}[1]
\algrenewcommand\algorithmicindent{1ex}
\algtext*{EndProcedure}
\algtext*{EndFor}
\algtext*{EndIf}

\Procedure{detect}{$\mathbf{S}_t$,$\mathcal{S}_\text{lost}$}
\For{$s \in \mathbf{S}_t$} \Comment{current segments}
\For{$\mathcal{O}_m \in \mathcal{S}_\text{lost}$} \Comment{inactive ``lost'' models}
\For{$\mathcal{K}_m \in \mathcal{O}_m$} \Comment{keypoint history}
    \State $ (\mathbf{T}, e) \leftarrow \text{RANSAC}(\mathcal{K}_m, \mathcal{K}_s)$ \Comment{estimation \labelcref{eq:est_transf_sparse}}
    \If{$e < 0.01$} 
        \State $\mathcal{S}_\text{lost} \leftarrow \mathcal{S}_\text{lost} \setminus \{\mathcal{O}_m\}$ \Comment{remove from lost set}
        \State $\mathcal{S} \leftarrow \mathcal{S} \cup \{\mathcal{O}_m\}$ \Comment{add to current set}
        \State $\mathbf{T}_m \leftarrow \mathbf{T}$ \Comment{reset object pose}
\EndIf 
\EndFor
\EndFor
\EndFor
\EndProcedure
\end{algorithmic}
\caption{Object redetection procedure.}
\label{algo:detect}
\end{algorithm}

\section{Evaluation}

\subsection{Setup}

\begin{figure}
    \centering
    \begin{subfigure}[b]{0.6\linewidth}
    \includegraphics[width=\linewidth]{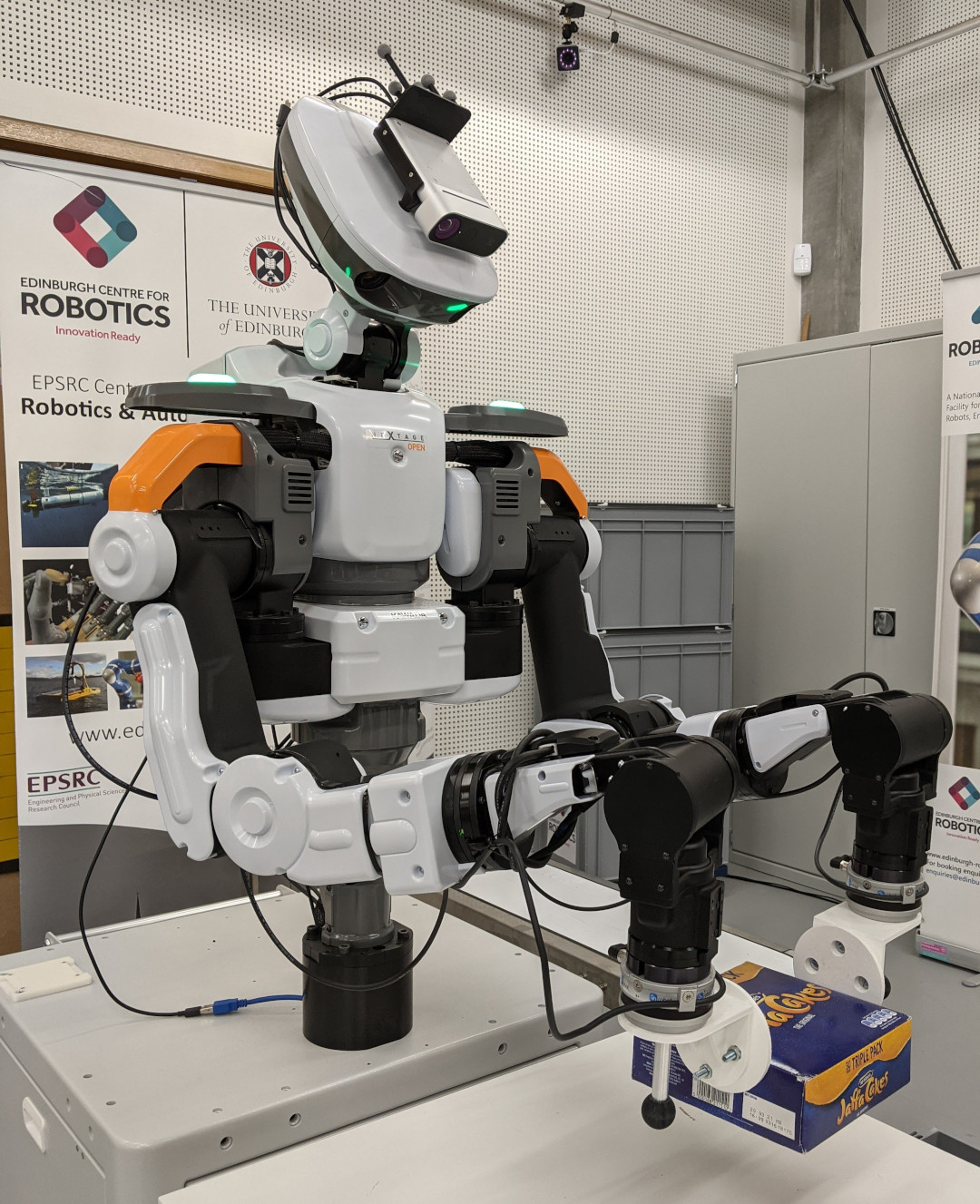}
    \caption{Nextage}
    \label{fig:nextage}
    \vspace{0.2\baselineskip}
    \end{subfigure}%
    \begin{subfigure}[b]{0.4\linewidth}
    \raisebox{1.5\height}{
    \begin{minipage}{\linewidth}
    \includegraphics[height=2.07cm]{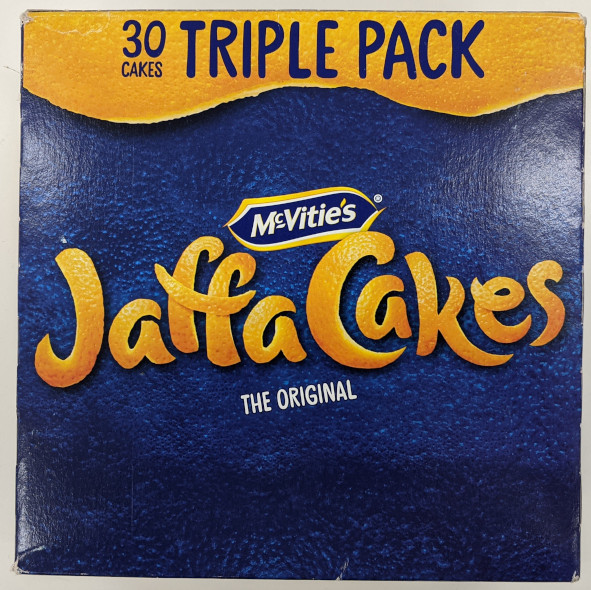}
    \includegraphics[height=2.07cm]{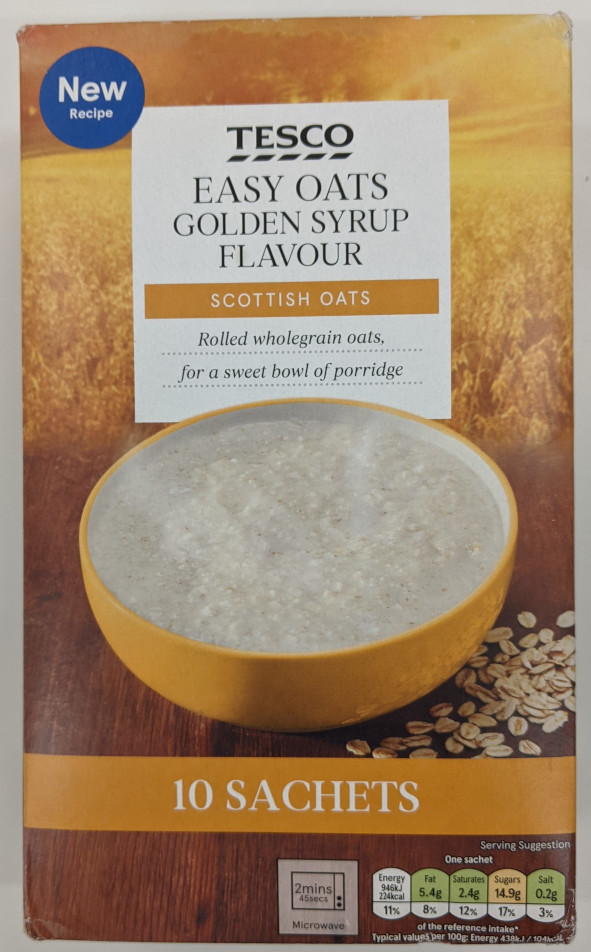}
    \includegraphics[width=\linewidth]{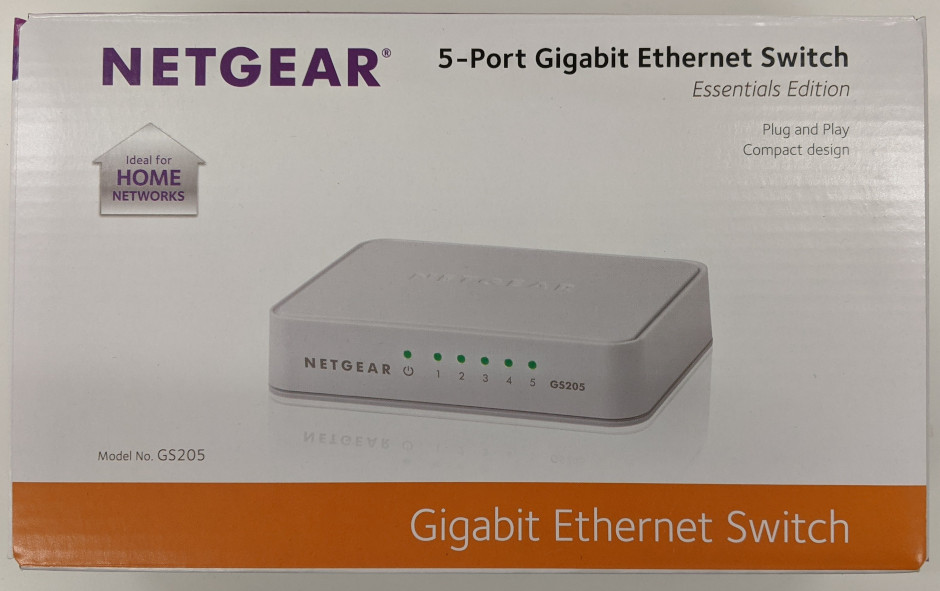}
    \end{minipage}
    }
    \caption{Objects}
    \label{fig:objects}
    \vspace{0.2\baselineskip}
    \end{subfigure}

    \caption{Experimental setup. (\subref{fig:nextage}) A stationary Nextage robot detects moving objects (\subref{fig:objects}) on a conveyor using a RGB-D camera mounted on the head, picks these objects from a conveyor using custom end-effectors and places them on a table. (\subref{fig:objects}) Objects from top left to bottom: \textit{jaffa}, \textit{oats}, \textit{netgear}.}
    \label{fig:nextage_setup}
\end{figure}

The proposed model-free tracking approach for grasping is evaluated on a Kawada Nextage robot (\Cref{fig:nextage_setup}). This robot is equipped with two 6-DoF arms fitted with custom end-effectors for dual-arm grasping. As RGB-D sensor, we use an Azure Kinect DK with a native resolution of $1280 \times 720$ after registering the depth to the colour frame. To reduce the computational costs and align the input image with the input size of the pretrained SuperPoint network, we crop and downscale the image to $640 \times 480$.

The ground-truth trajectory for the camera and object motion is provided by a Vicon system using markers attached to the camera body and to the conveyor respectively. The trajectory estimation error is quantified via the absolute trajectory error (ATE) and the relative pose error (RPE) \cite{Sturm2012}. The ground-truth environment reconstruction is created from the ground-truth camera trajectory and the reconstruction error is quantified by the point distances between true and estimated reconstruction. The robot links are not considered as moving objects and filtered from the depth data.

\subsection{Transformation Estimation}

In a typical manipulation task, a robot has to change its FoV between the pick and place targets. The stationary Nextage is only capable to change the FoV by rotating the torso or head. This rotation motion is especially challenging for dense ICP approaches since the overlap between consecutive images is small. We selected two sequences to evaluate our approach \emph{MultiMotionFusion} (MMF). In the \textit{manipulation} sequence, the robot rotates its FoV multiple times between different points on the conveyor belt and the table. In the \textit{rotation} sequence the torso rotates two half rotations ($180\,\text{deg}$) forth and back in one go. The torso always rotates at maximum speed, albeit the \textit{rotation} sequence will have a higher peak speed and is therefore more challenging.

We compare the proposed approach (MMF) to ElasticFusion (EF \cite{Whelan2015}), StaticFusion (SF \cite{Scona2018}), RigidFusion (RF \cite{Long2021}) and Co-Fusion (CF \cite{Ruenz2017}). Since estimation errors directly lead to segmentation errors, we also run CF without a segmentation using a single model. To investigate the benefit of additional dense refinement in our approach, we compare the sparse keypoint-only approach with the full proposed pipeline that additionally does a dense refinement.
The absolute trajectory error (ATE) and the relative pose error (RPE) in \Cref{tab:kpinit_errors} show that dense ICP methods (EF, SF, CF) fail to track the fast rotation motion of the camera. The proposed sparse keypoint-only approach (\emph{MMF (sparse)}) prevents these failures and the additional dense refinement (\emph{MMF (s+d)}) further improves the tracking results. RF uses the robot kinematic as motion prior and therefore performs much better than pure dense approaches and also achieves the lowest rotational errors on the \textit{manipulation} sequence.

\begin{table}
\setlength\tabcolsep{4pt}
\begin{subtable}{\linewidth}
\centering
\begin{tabular}{|c|r|r|r|r|r|r|r|}
\hline
\textbf{seq.} &
  \multicolumn{1}{c|}{\textbf{EF}} &
  \multicolumn{1}{c|}{\textbf{SF}} &
  \multicolumn{1}{c|}{\textbf{RF}} &
  \multicolumn{1}{c|}{\textbf{CF}} &
  \multicolumn{1}{c|}{\textbf{\begin{tabular}[c]{@{}c@{}}CF\\ (static)\end{tabular}}} &
  \multicolumn{1}{c|}{\textbf{\begin{tabular}[c]{@{}c@{}}MMF\\ (sparse)\end{tabular}}} &
  \multicolumn{1}{c|}{\textbf{\begin{tabular}[c]{@{}c@{}}MMF\\ (s+d)\end{tabular}}} \\ \hline
\textit{manip.} &
  67.03 &
  81.83 &
  1.83 &
  63.26 &
  30.35 &
  3.66 &
  \textbf{1.68} \\ \hline
\textit{rotation} &
  49.04 &
  73.18 &
  8.80 &
  59.79 &
  42.77 &
  7.79 &
  \textbf{2.25} \\ \hline
\end{tabular}
\caption{Transl. ATE RMSE (cm)}
\vspace{0.5\baselineskip}
\end{subtable}

\vspace{6pt}

\begin{subtable}{\linewidth}
\centering
\centering
\begin{tabular}{|c|r|r|r|r|r|r|r|}
\hline
\textbf{seq.} &
  \multicolumn{1}{c|}{\textbf{EF}} &
  \multicolumn{1}{c|}{\textbf{SF}} &
  \multicolumn{1}{c|}{\textbf{RF}} &
  \multicolumn{1}{c|}{\textbf{CF}} &
  \multicolumn{1}{c|}{\textbf{\begin{tabular}[c]{@{}c@{}}CF\\ (static)\end{tabular}}} &
  \multicolumn{1}{c|}{\textbf{\begin{tabular}[c]{@{}c@{}}MMF\\ (sparse)\end{tabular}}} &
  \multicolumn{1}{c|}{\textbf{\begin{tabular}[c]{@{}c@{}}MMF\\ (s+d)\end{tabular}}} \\ \hline
\textit{manip.} &
  101.21 &
  134.76 &
  \textbf{2.72} &
  110.75 &
  61.78 &
  5.45 &
  3.04 \\ \hline
\textit{rotation} &
  90.41 &
  120.17 &
  12.60 &
  111.15 &
  86.36 &
  11.59 &
  \textbf{4.78} \\ \hline
\end{tabular}
\caption{Transl. RPE RMSE (cm/s)}
\vspace{0.5\baselineskip}
\end{subtable}

\vspace{6pt}

\begin{subtable}{\linewidth}
\centering
\begin{tabular}{|c|r|r|r|r|r|r|r|}
\hline
\textbf{seq.} &
  \multicolumn{1}{c|}{\textbf{EF}} &
  \multicolumn{1}{c|}{\textbf{SF}} &
  \multicolumn{1}{c|}{\textbf{RF}} &
  \multicolumn{1}{c|}{\textbf{CF}} &
  \multicolumn{1}{c|}{\textbf{\begin{tabular}[c]{@{}c@{}}CF\\ (static)\end{tabular}}} &
  \multicolumn{1}{c|}{\textbf{\begin{tabular}[c]{@{}c@{}}MMF\\ (sparse)\end{tabular}}} &
  \multicolumn{1}{c|}{\textbf{\begin{tabular}[c]{@{}c@{}}MMF\\ (s+d)\end{tabular}}} \\ \hline
\textit{manip.} &
  33.13 &
  40.72 &
  \textbf{1.94} &
  44.66 &
  21.00 &
  2.40 &
  2.50 \\ \hline
\textit{rotation} &
  39.50 &
  51.51 &
  3.74 &
  65.74 &
  47.88 &
  4.12 &
  \textbf{3.60} \\ \hline
\end{tabular}
\caption{Rotat. RPE RMSE (deg/s)}
\vspace{0.5\baselineskip}
\end{subtable}

\caption{ATE and RPE for \textit{manipulation} and \textit{rotation} sequences. Dense approaches (EF, SF, CF) fail due to the high camera motion. The proposed sparse approach (MMF) can handle this motion and further improve the translational errors using the dense refinement. The refinement slightly degrades the rotational alignment.}
\label{tab:kpinit_errors}
\end{table}

\begin{figure*}
    \centering
    \setlength\tabcolsep{0pt}
    \renewcommand{\arraystretch}{0}
    \newcommand*\rot[1]{\rotatebox[origin=c,y=2.2cm]{90}{#1}}
    \begin{tabular}{rcccc}
     &
      \textbf{ground truth} &
      \textbf{CF (static)} &
      \textbf{MMF (sparse)} &
      \textbf{MMF (s+d)} \vspace{2pt} \\
    \rot{\textit{manipulation}} \hspace{1pt} &
      \includegraphics[width=0.24\textwidth,frame]{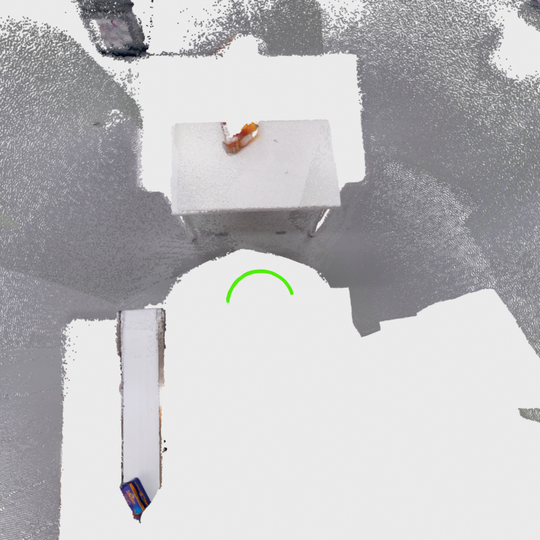} &
      \includegraphics[width=0.24\linewidth,frame]{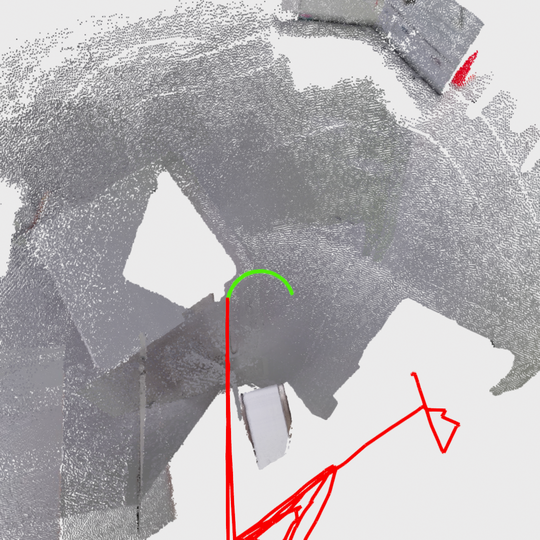} &
      \includegraphics[width=0.24\linewidth,frame]{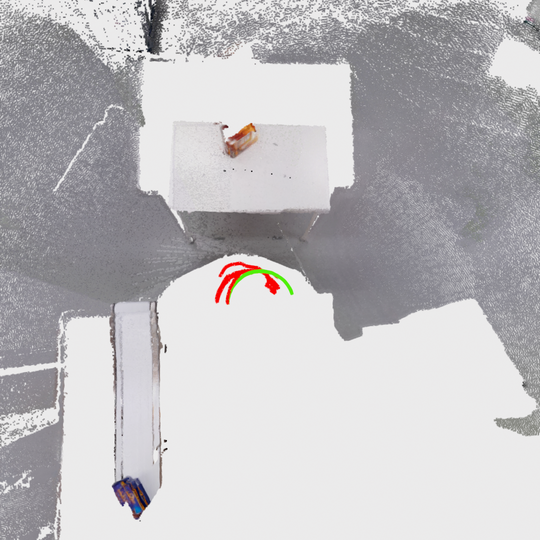} &
      \includegraphics[width=0.24\linewidth,frame]{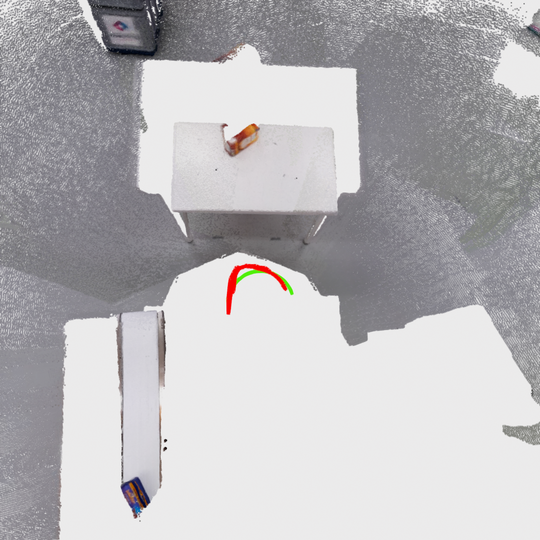} \\
    \rot{\textit{rotation}} \hspace{1pt} &
      \includegraphics[width=0.24\linewidth,frame]{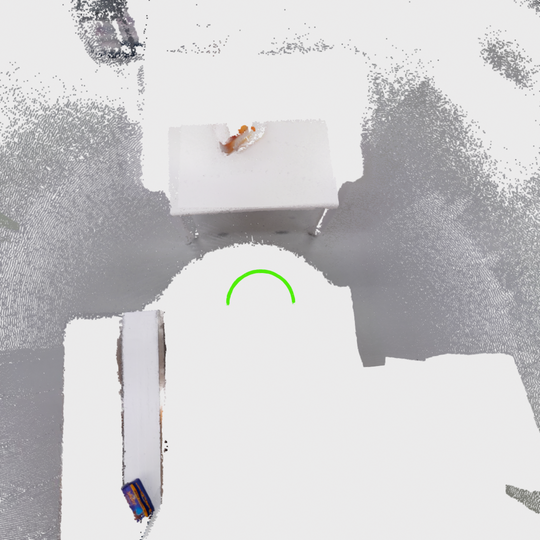} &
      \includegraphics[width=0.24\linewidth,frame]{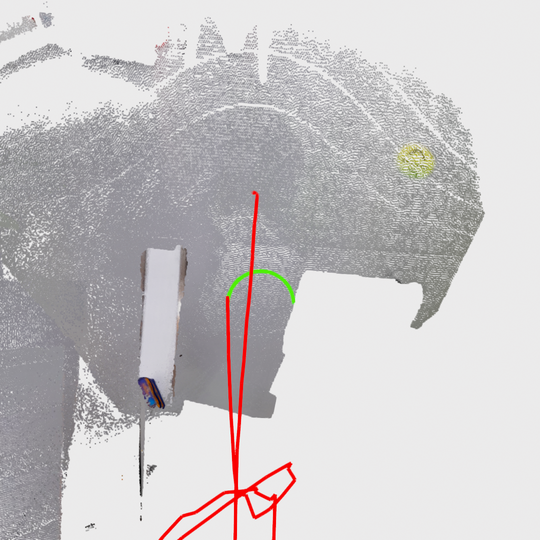} &
      \includegraphics[width=0.24\linewidth,frame]{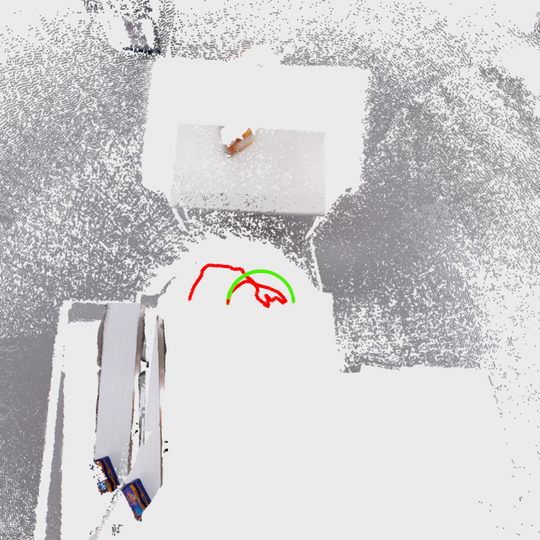} &
      \includegraphics[width=0.24\linewidth,frame]{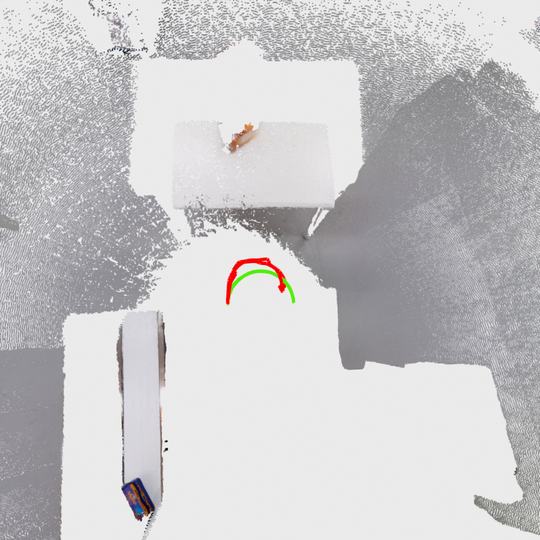} \\
    \end{tabular}
    \caption{Estimated (red) and true (green) camera trajectory with resulting reconstruction of the environment for two sequences. In our approach (MMF) the keypoints \emph{(sparse)} prevent a failure of the point cloud alignment, while the dense refinement \emph{(s+d)} prevents drift.}
    \label{fig:kpinit_reconstr}
\end{figure*}

The qualitative comparison of the estimated camera trajectory and the environment reconstruction in \Cref{fig:kpinit_reconstr} visualises that the keypoint estimation mostly keeps the conveyor aligned with the table, while the dense refinement further improves the alignment in consecutive frames.

\begin{figure*}
    \centering
    \includegraphics[width=0.33\linewidth,frame]{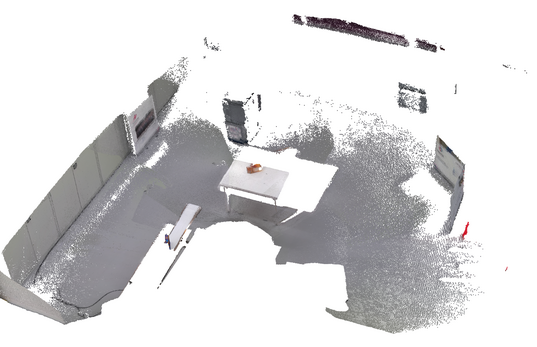}%
    \includegraphics[width=0.33\linewidth,frame]{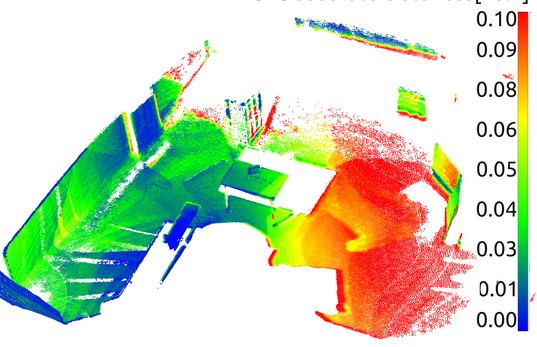}%
    \includegraphics[width=0.33\linewidth,frame]{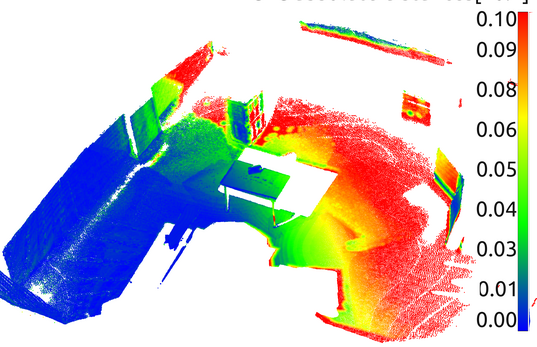}
    \caption{Point cloud reconstruction error for \textit{manipulation}, \textbf{left:} reference via ground truth camera trajectory, \textbf{centre:} from sparse keypoints only (MSE: $3.81\pm2.97\,$cm), \textbf{right:} with dense refinement (MSE: $2.81\pm3.35\,$cm). The dense refinement of the sparse keypoint estimate improves the reconstruction specifically when facing the conveyor on the left side again.}
    \label{fig:kpinit_reconstr_error}
\end{figure*}

Finally, the quantitative comparison of the reconstructed environment models in \Cref{fig:kpinit_reconstr_error} further demonstrates that the sparse estimation coarsely aligns the point clouds, while the dense refinement mostly improves the alignment in the vicinity of the robot.

\subsection{Motion Segmentation}

The object tracking is evaluated separately from the camera tracking using three different objects (\Cref{fig:objects}) that are moving on a conveyor belt at $6.8\,\text{cm/s}$. We compare the tracking results of the proposed approach MMF against RigidFusion (RF) and Co-Fusion (CF) in two configurations. The box-shaped objects can either stand \textit{up} with the second largest side orthogonal to the ground, or laying flat \textit{down} with the smallest side orthogonal to the ground.

For grasping, we are primarily interested in consistently tracking a given frame in the object model. This reference frame can be chosen arbitrarily and is typically set to the camera frame at the point in time when this object is first detected. For visualisation purposes, we set this frame to the centre of the object segment when it is first detected. 

\begin{figure}
    \centering
    \setlength\tabcolsep{0pt}
    \newcommand\xl{0.1} 
    \newcommand\xu{0.4} 
    \renewcommand{\arraystretch}{0}
    \newcommand*\rot[1]{\rotatebox[origin=c,y=1cm]{90}{#1}}
    \begin{tabular}{rcc}
     & \textit{jaffa} & \textit{oats} \vspace{2pt} \\
    \rot{\textbf{RF}} \hspace{0.1pt} &
        \includegraphics[width=0.47\linewidth,frame]{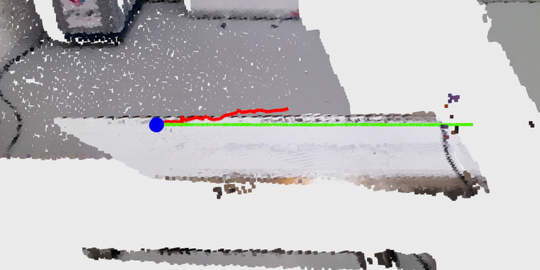} &
        \includegraphics[width=0.47\linewidth,frame]{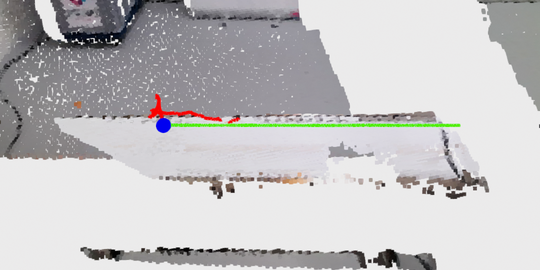}\\
    \rot{\textbf{CF}} \hspace{0.1pt} &
        \includegraphics[width=0.47\linewidth,frame]{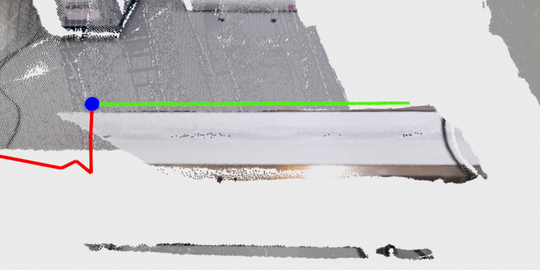} &
        \includegraphics[width=0.47\linewidth,frame]{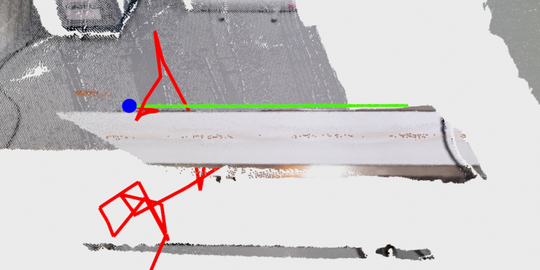}\\
    \rot{\textbf{MMF}} \hspace{0.1pt} &
        \includegraphics[width=0.47\linewidth,frame]{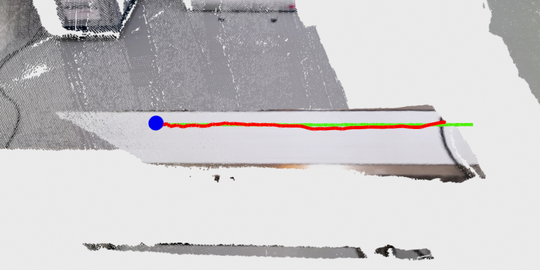} &
        \includegraphics[width=0.47\linewidth,frame]{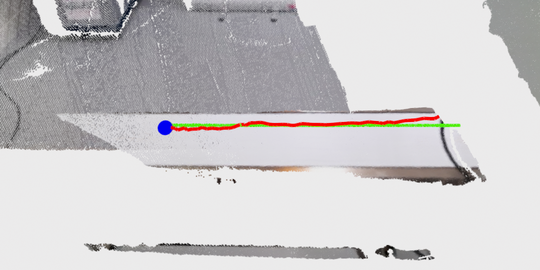}\\
    \end{tabular}
    \caption{Estimated (red) and true (green) object trajectory on conveyor belt from the point where an object's segment centre is first detected (blue). The instant motion segmentation in MMF leads to a much more consistent tracking of the reference frame.}
    \label{fig:segm_tracking}
\end{figure}

The proposed approach provides a much more consistent tracking of this reference frame (\Cref{fig:segm_tracking}) and also produces a lower ATE (\Cref{tab:segm_ate}) than the baseline approaches. The CF segmentation via the dense reprojection error fails to detect any object laying \textit{down} flat on the conveyor.

\begin{table}
\centering
\begin{tabular}{|c|lrrr|}
\hline
\textbf{seq.} &
  \multicolumn{1}{c|}{\textbf{type}} &
  \multicolumn{1}{c|}{\textbf{RF}} &
  \multicolumn{1}{c|}{\textbf{CF}} &
  \multicolumn{1}{c|}{\textbf{MMF}} \\ \hline
\multirow{2}{*}{\textit{jaffa}}                         & up   & 16.04 & 37.50 & \textbf{1.31} \\
                                                        & down & 17.43 & ---   & \textbf{1.02} \\ \hline
\multirow{2}{*}{\textit{oats}}                          & up   & 14.94 & 31.48 & \textbf{1.02} \\
                                                        & down & 17.99 & ---   & \textbf{1.19} \\ \hline
\multicolumn{1}{|l|}{\multirow{2}{*}{\textit{netgear}}} & up   & 19.88 & 82.69 & \textbf{1.17} \\
                                                        & down & 21.30 & ---   & \textbf{1.02} \\ \hline
\end{tabular}
\caption{Transl. ATE (cm) for tracking the object centre from where they are initially detected up to the grasping position. A dash indicates that no object was detected.}
\label{tab:segm_ate}
\end{table}

\begin{figure}
    \centering
    \setlength\tabcolsep{0pt}
    \renewcommand{\arraystretch}{0}
    \newcommand*\rot[1]{\rotatebox[origin=c,y=1cm]{90}{#1}}

    \begin{subtable}{\linewidth}
    \centering
    \begin{tabular}{rccc}
     & $10.6\,\text{s}$ & $15.7\,\text{s}$ & $18.9\,\text{s}$ \vspace{2pt} \\
     \rot{\textbf{CF}} \hspace{0.1pt} &
        \includegraphics[width=0.3\linewidth,frame]{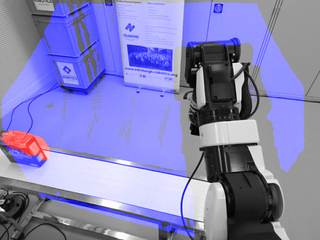} &
        \includegraphics[width=0.3\linewidth,frame]{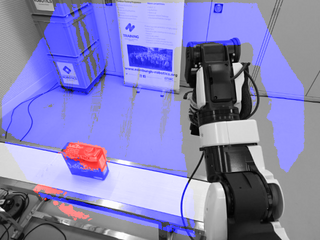} &
        \includegraphics[width=0.3\linewidth,frame]{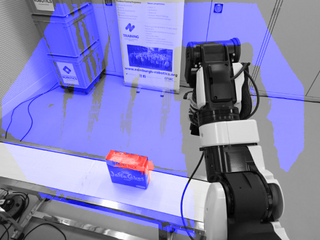} \\
    \rot{\textbf{MMF}} \hspace{0.1pt} &
        \includegraphics[width=0.3\linewidth,frame]{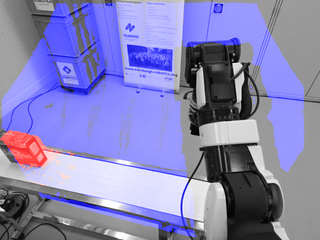} &
        \includegraphics[width=0.3\linewidth,frame]{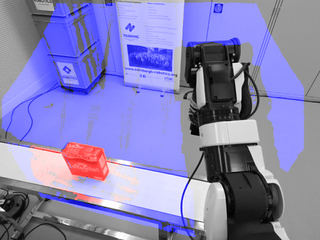} &
        \includegraphics[width=0.3\linewidth,frame]{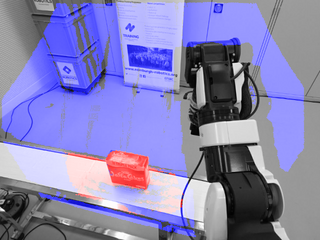} \\
    \end{tabular}
    \caption{\textit{jaffa}}
    \vspace{0.5\baselineskip}
    \end{subtable}

    \vspace{6pt}
    
    \begin{subtable}{\linewidth}
    \centering
    \begin{tabular}{rccc}
     & $10.9\,\text{s}$ & $15.9\,\text{s}$ & $19.6\,\text{s}$ \vspace{2pt} \\
     \rot{\textbf{CF}} \hspace{0.1pt} &
        \includegraphics[width=0.3\linewidth,frame]{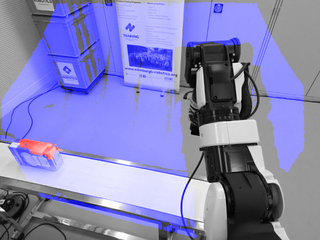} &
        \includegraphics[width=0.3\linewidth,frame]{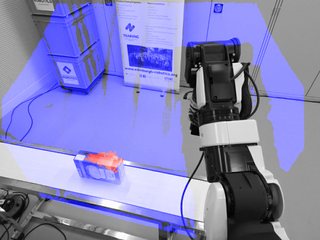} &
        \includegraphics[width=0.3\linewidth,frame]{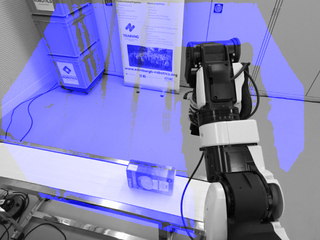} \\
    \rot{\textbf{MMF}} \hspace{0.1pt} &
        \includegraphics[width=0.3\linewidth,frame]{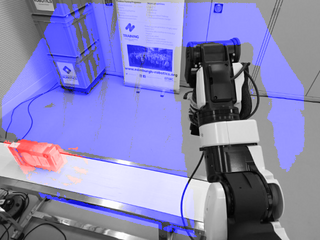} &
        \includegraphics[width=0.3\linewidth,frame]{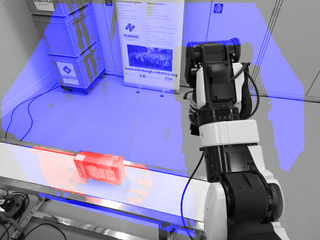} &
        \includegraphics[width=0.3\linewidth,frame]{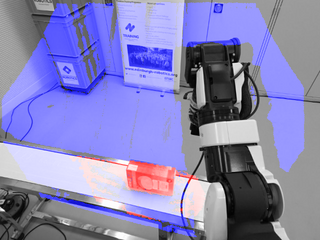} \\
    \end{tabular}
    \caption{\textit{oats}}
    \vspace{0.5\baselineskip}
    \end{subtable}
    
    \caption{Segmentation of the environment (blue) and the object (red) blended with the original image. The baseline (CF) looses track of the object towards the end of the sequence.}
    \label{fig:segm_overlay}
\end{figure}

This improved and consistent tracking is achieved by segmenting the object in one instance (\Cref{fig:segm_overlay}). The segmentation via the raw dense reprojection error only signifies motion where a large error is observed, such as between the top of the object and the ground plane, and thus never segments the entire object. The proposed keypoint and optical flow segmentation creates a larger motion segment, covering the entire object, and thus provides instantly more data to align consecutive frames using keypoint and dense data.

\begin{figure}
    \centering
    \includegraphics[width=0.33\linewidth,frame]{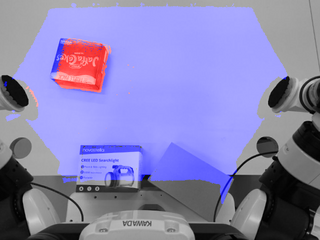}%
    \includegraphics[width=0.33\linewidth,frame]{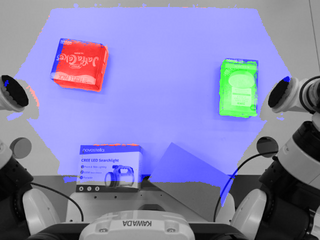}%
    \includegraphics[width=0.33\linewidth,frame]{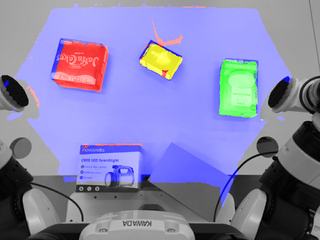}%
    \caption{Segmentation of multiple static objects (red, green, yellow) that are slid by a human on a table one-by-one.}
    \label{fig:segm_multi_object}
\end{figure}

While our approach requires an initial motion, it is capable of segmenting multiple objects with irregular motions (\Cref{fig:segm_multi_object}) and continues tracking after the motion stopped.

\subsection{Model Redetection and Grasping}

A manipulation experiment applies the tracking, segmentation and redetection to a dual-arm grasping task to demonstrate their feasibility to continuously track and interact with previously unknown objects over a longer time period.

\begin{figure}
    \centering
    \newcommand\w{123.48} 
    \begin{picture}(\w,\w)
    \put(0,0){\includegraphics[width=\w pt,frame]{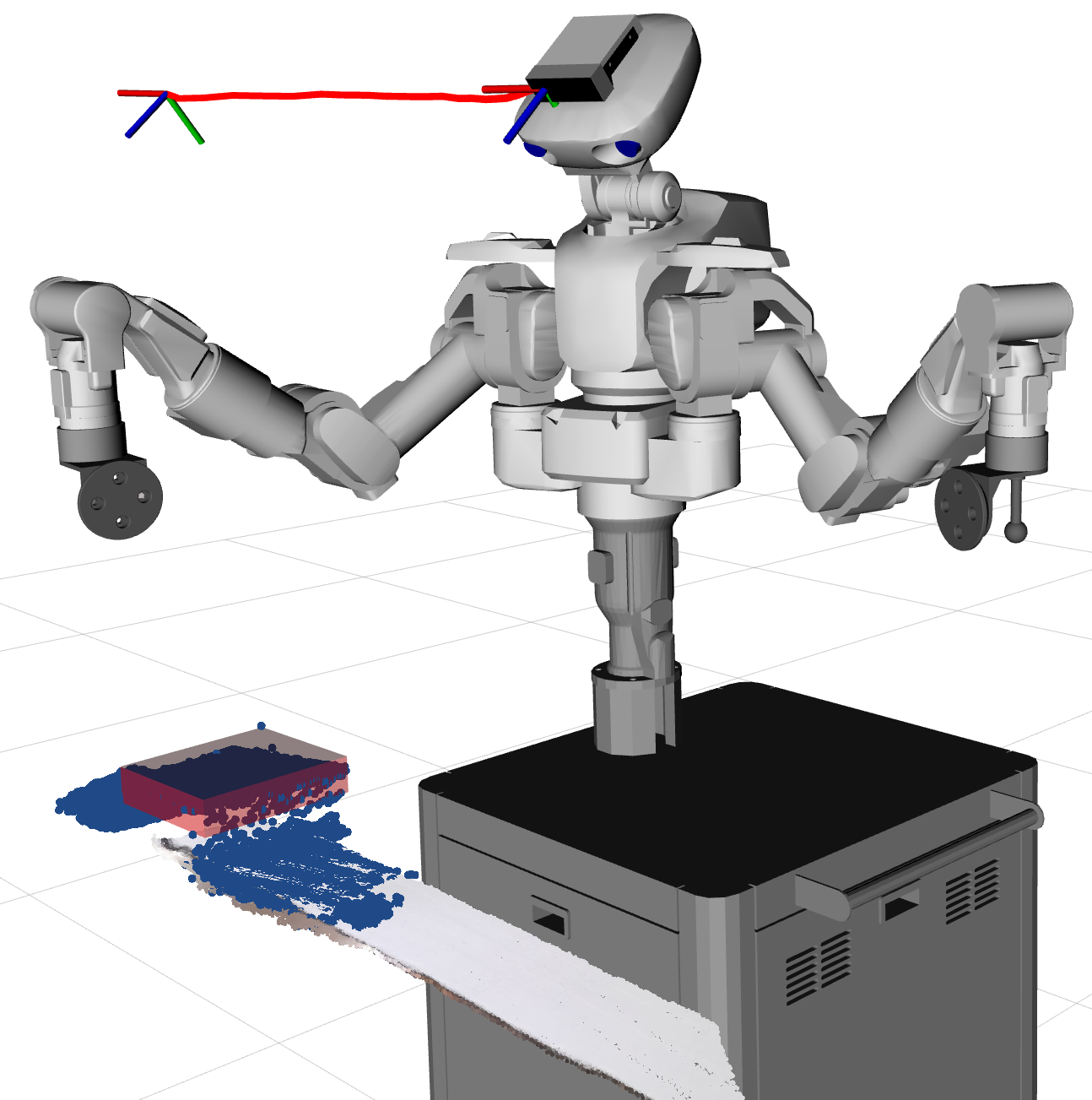}}
    \put(100,105){\circled{1}}
    \end{picture}%
    \begin{picture}(\w,\w)
    \put(0,0){\includegraphics[width=\w pt,frame]{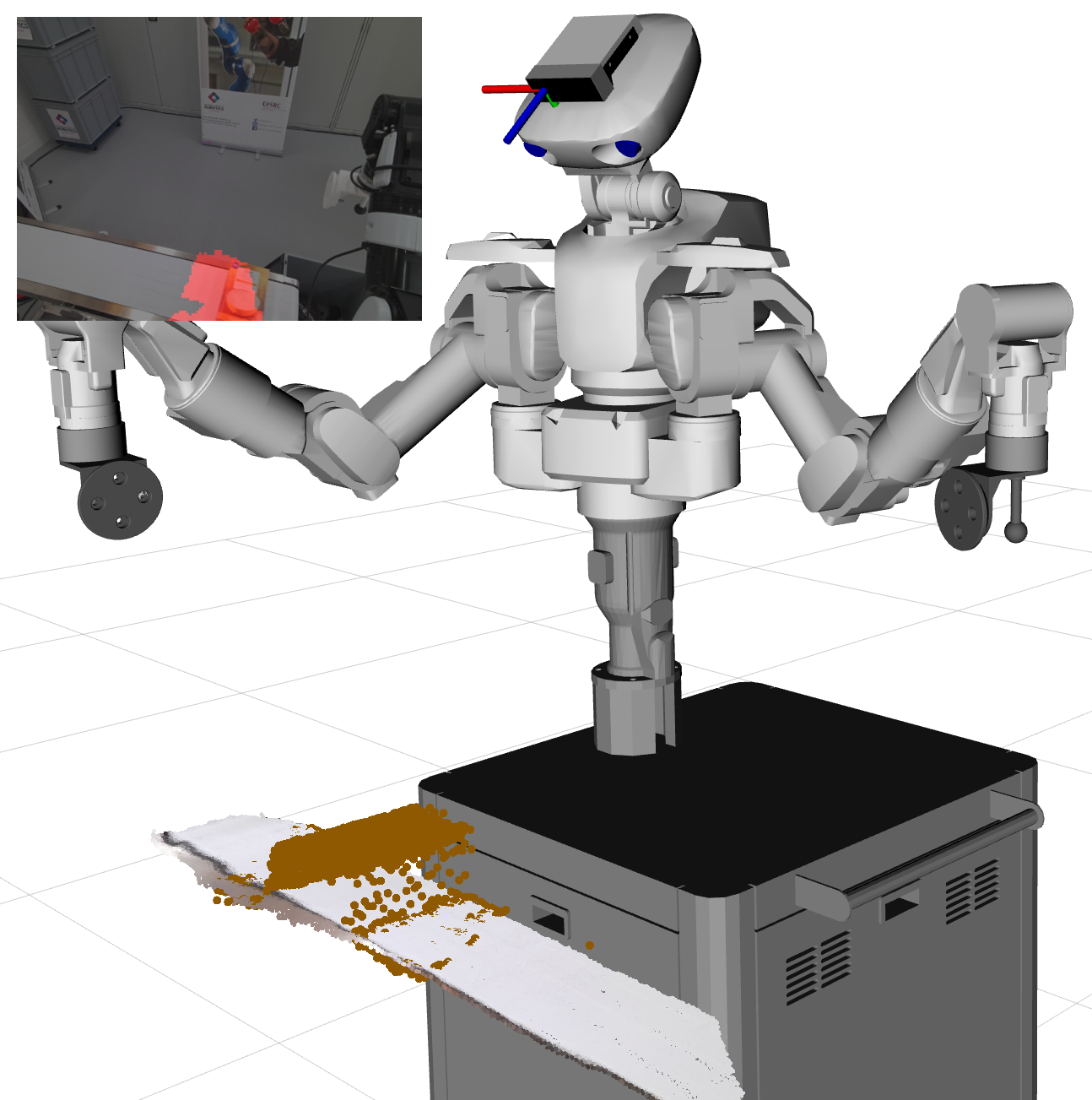}}
    \put(100,105){\circled{2}}
    \end{picture}
    
    \begin{picture}(\w,\w)
    \put(0,0){\includegraphics[width=\w pt,frame]{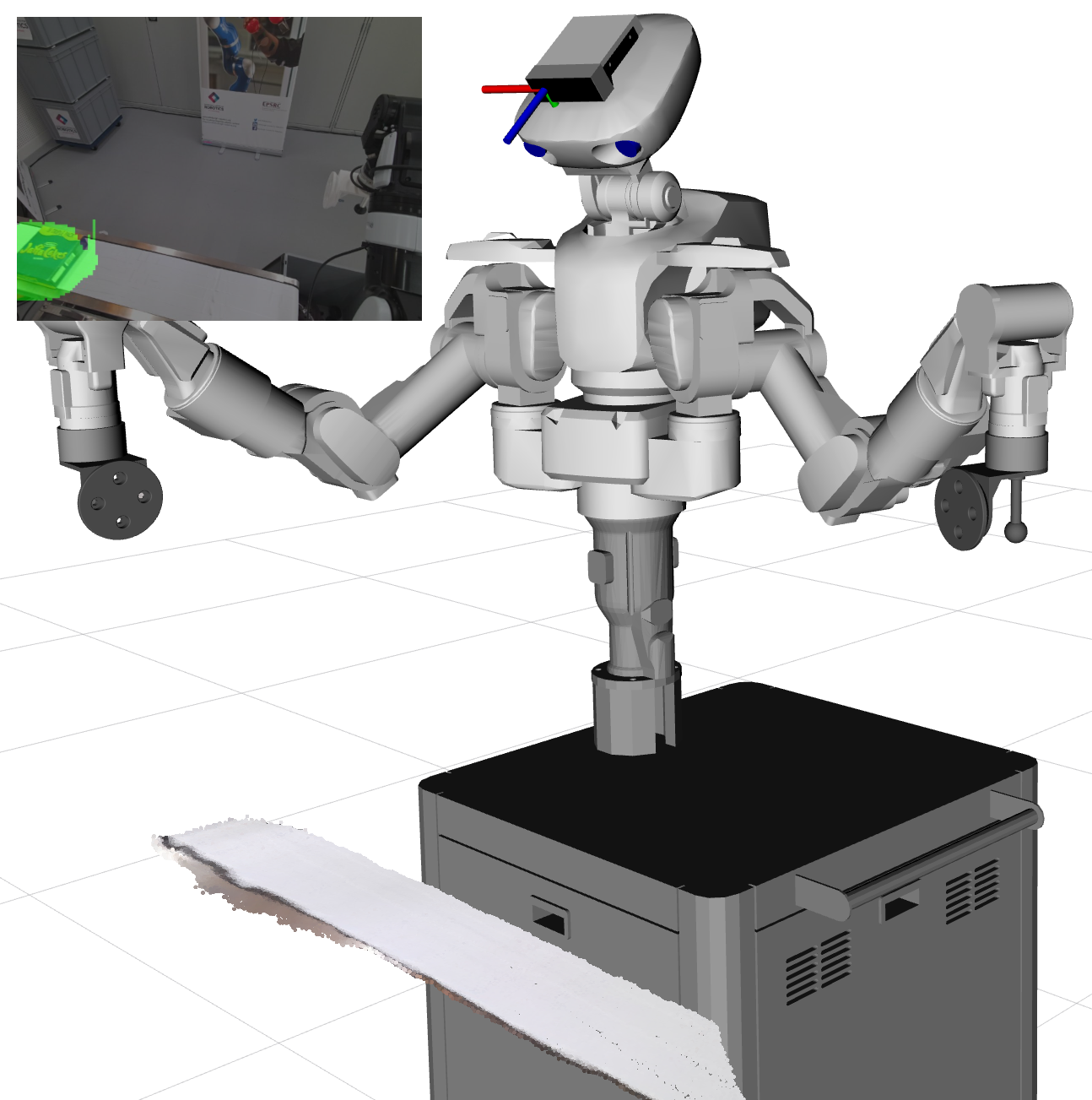}}
    \put(100,105){\circled{3}}
    \end{picture}%
    \begin{picture}(\w,\w)
    \put(0,0){\includegraphics[width=\w pt,frame]{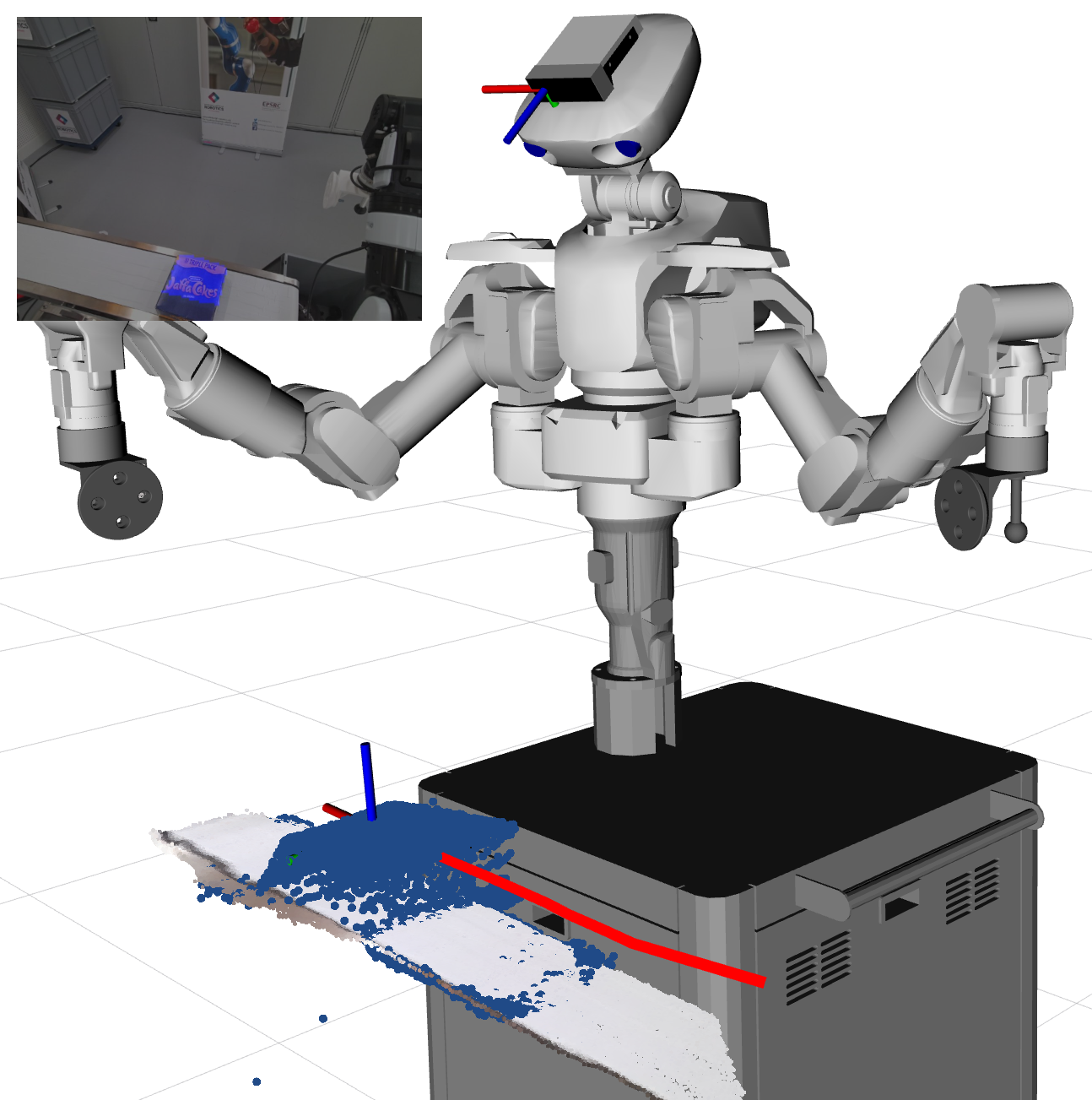}}
    \put(100,105){\circled{4}}
    \end{picture}
    
    \caption{Phases of online modelling new objects and redetecting previous objects. The robot observes consecutive objects on the conveyor, models their dense and sparse representation and replaces newly created objects with previous models if they match. Together with the redetected model, we recall a stored bounding-box frame as grasping target.}
    \label{fig:pnp_redetect}
\end{figure}

In the sequence (\Cref{fig:pnp_redetect}), the camera is facing the moving conveyor ($19.5\,\text{cm/s}$) and we initially place the \textit{jaffa} $\circled{1}$ and the \textit{oats} $\circled{2}$ object on the conveyor. When moving in the FoV, we extract an oriented bounding-box on the modelled dense point cloud to extract the centre and width of the \textit{jaffa} object. The calibration of this new frame is stored as the new grasp reference frame. We then place the \textit{jaffa} box again on the conveyor. Initially, this will be seen as a third object (green segment $\circled{3}$). Shortly after, this segment is correctly matched with the previous \textit{jaffa} model and replaced (blue segment $\circled{4}$). With this restored model, we can also restore the grasp reference frame in the object centre which is then tracked and grasped once the object stops. See the supplementary video for details.

\subsection{Runtime}

The evaluation runs on an Intel Core i9-9900KF with a Nvidia GeForce RTX 2080 SUPER.
The average runtime of the individual stages are as follows: keypoint extraction and matching, 18\,ms and 17\,ms respectively; sparse and dense transformation estimation, 17\,ms; optical flow, 9\,ms; CRF segmentation, 43\,ms; re-detection, 2.3\,ms. In total with other minor stages, the full pipeline takes 126\,ms per image (8\,Hz).


\section{Conclusion}

This work motivated the use of model-free object tracking approaches for robotic manipulation tasks, to overcome limitations with model-based approaches and their biased dataset selection where the scene and the manipulated objects are not known in advance.
We further argued and demonstrated that a direct motion estimation via sparse keypoints provides a much more robust transformation estimation and segmentation in comparison to indirect motion inference from ambiguous dense data.
The combination of a sparse and dense model representation enables robust tracking and segmentation of previously unseen objects, and thus enables robotic manipulation tasks without prior object models.

The keypoint association and the optical flow propagation are the critical parts of our pipeline. While SuperPoints are robust to large displacements, the descriptors are limited to about $45\,\text{deg}$ in-plane rotation which restricts the redetection. Spurious keypoints have a negative impact on the unary CRF potentials, resulting in ``flooding'' of segments into nearby areas and thus an overestimation of the segment size.
In future work, we would like to combine the keypoint correspondence and optical flow tasks to relate pixels over short and long distances and mitigate some of these effects.


\bibliographystyle{IEEEtran}
\bibliography{references}

\end{document}